\documentclass[10pt,twocolumn,letterpaper]{article}

\usepackage{wacv}
\usepackage{times}
\usepackage{epsfig}
\usepackage{graphicx}
\usepackage{amsmath}
\usepackage{amssymb}
\usepackage{multirow}
\usepackage{enumitem}
\usepackage{textgreek}
\usepackage{color}

\def\wacvPaperID{} 

\wacvfinalcopy 

\ifwacvfinal
\fi

\ifwacvfinal
\usepackage[breaklinks=true,bookmarks=false]{hyperref}
\else
\usepackage[pagebackref=true,breaklinks=true,colorlinks,bookmarks=false]{hyperref}
\fi

\title{Weakly Supervised Annotations for Multi-modal Greeting Cards Dataset}  

\author{Sidra Hanif\\
Temple University, PA\\
{\tt\small sidra.hanif@temple.edu}
\and
Longin Jan Latecki\\
Temple University, PA\\
{\tt\small latecki@temple.edu}
}
\begin{document}
\maketitle



\begin{abstract}
In recent years, there is a growing number of pre-trained models trained on a large corpus of data and yielding good performance on various tasks such as classifying multimodal datasets. These models have shown good performance on natural images but are not fully explored for scarce abstract concepts in images. In this work, we introduce an image/text-based dataset called Greeting Cards.
Dataset (GCD) that has abstract visual concepts. In our work, we propose to aggregate features from pretrained images and text embeddings to learn abstract visual concepts
from GCD. This allows us to learn the text-modified image features, which combine complementary and redundant information from the multi-modal data streams into a single, meaningful feature.
Secondly, the captions for the GCD dataset are computed with the pretrained CLIP-based image captioning model. Finally, we also demonstrate that the proposed
the dataset is also useful for generating greeting card images using pre-trained text-to-image generation model.
\end{abstract}
\begin{figure}
\begin{center}
\includegraphics[scale=0.4]{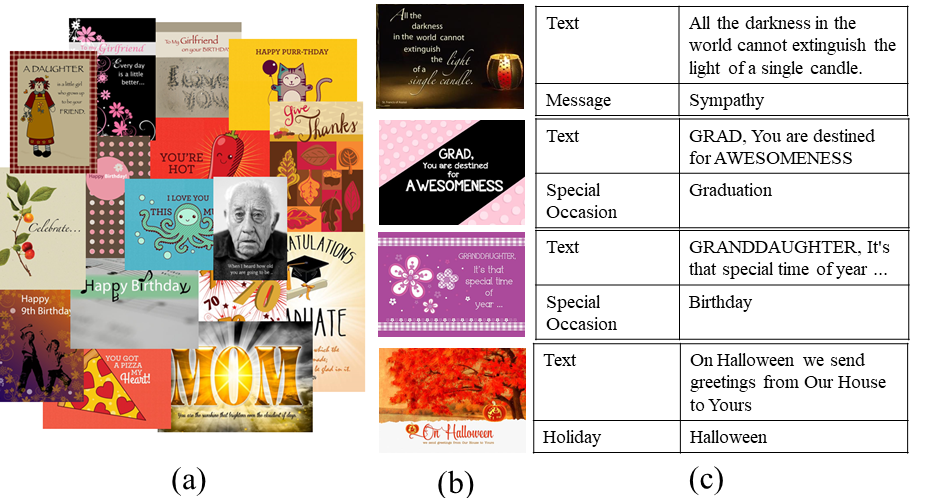}
   
\caption{Sample images from datasets (b, c) Text and labels in GCD for labels in categories Holidays, Special Occasions and Messages.}
\label{fig:sample_GCD}
\end{center} 
\end{figure}

\begin{figure}
\begin{center}
\includegraphics[scale=0.34]{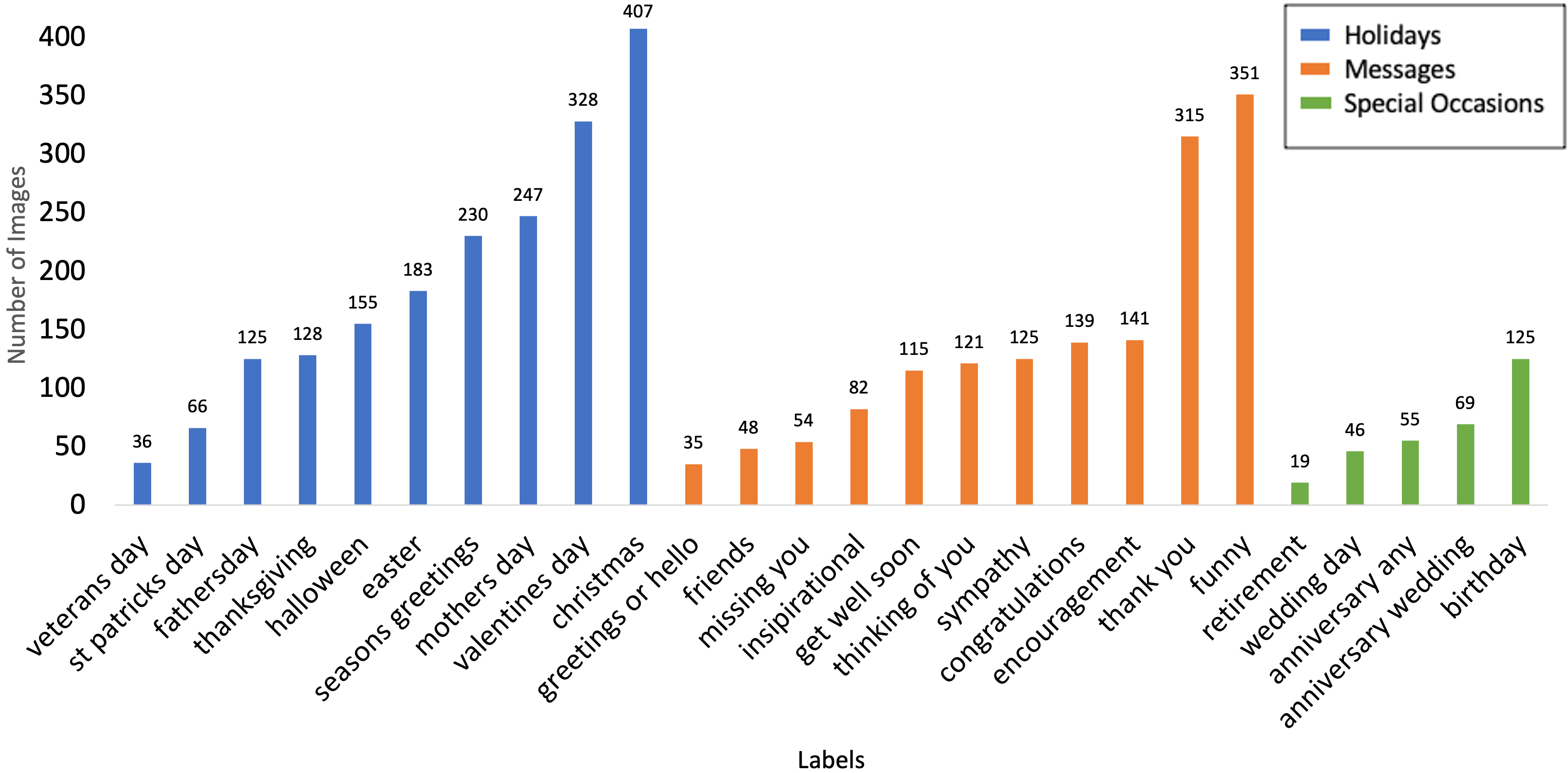}
   
\caption{The number of images in each category for Holidays, Messages and Special occasions.}
\label{fig:num_images}
\end{center} 
\end{figure}


\section{Introduction}


Nowadays, it is a common practice in computer vision to consider multimodal information, as in the real world, information usually comes in multiple modalities. The most common and information rich modalities are images and text. For example, images are usually associated with textual explanations. 
Considering the importance of multimodal datasets, several multimodal datasets originated almost decades back with the introduction of MS-COCO~\cite{lin2014microsoft} and Flickr30K~\cite{plummer2015flickr30k}. A large scale image-text dataset such as SBU Captions~\cite{ordonez2011im2text}, CC~\cite{changpinyo2021cc12m} and visual genome \cite{krishna2017visual} datasets are the center of attention for research community.
Despite of this bulk of multimodal datasets, almost all datasets have only one image-text pair. 
The exception to this trend is the recent dataset WIT~\cite{srinivasan2021wit}, an  image-text datasets with descriptive labels through the form of attribute text and titles. The datasets in WIT is downloaded from Wikipedia, which uses crowd-sourcing in the data creation process. In this dataset images are accompanied by illustrations and detailed text descriptions (Wikipedia).


Different from the previous datasets, we developed a dataset called the Greeting Cards Datasets (GCD). The dataset is a subset of greeting cards provided by an online greeting card company SignedCards.com \cite{signedcards}. Each greeting card has three entities of information, i.e., cover image (image), cover text (text) and category (abstract label), so the data is in a format of image-text-abstract label. In terms of images, a cover photo for each card in the dataset is available. Most of the images are human generated (e.g., paintings, caricature, clip art, cartoons and sketches). For the computer vision domain, GCD provides a data source for researches in non-natural images, conveying the main idea of the greeting cards in terms of abstract concepts. For textual information, the cover text for each card in the dataset is extracted. In terms of label, there are three categories namely Holidays, Special occasion, and Messages. The text is manually extracted from cover image and abstract label is assigned to each card by the engineer at \cite{signedcards}.
Sample cards with their related images, texts and labels are shown in Figure~\ref{fig:sample_GCD}(a,b,c).


From the computer vision perspective, our data offers unique conceptual information but it is also challenging to develop computer vision algorithms such as classification or retrieval algorithms of greeting cards. It is different from existing datasets from the point of view of computer vision in two ways. First, most of the visual objects' appearance is unique and have no similar look in other images in the dataset. It contains large amount of cartoonistic and synthesized images. 
Sample images are shown in Figure~\ref{fig:sample_GCD}(a).
Such kinds of images are widely available on the web, but they have not been properly explored in computer vision domain. This dataset has extraordinary intra-class variance.  
Second, unlike the popular computer vision datasets, our dataset is sparse. The number of images in each category is also imbalanced as seen in Figure~\ref{fig:num_images}. Due to above mentioned challenges, we expect regular supervised algorithms, which require training, to have hard times to learn a general representation on this abstract and sparse dataset. 


Since GCD is a small-scale datasets, pretrained models are one of the feasible choices to compute generic feature representation. Recently proposed multimodal visio-linguistic models, such as CLIP~\cite{radford_2021_learning} with superior performance was trained on billions of image-text pairs. Cutting-edge language model like Google's Universal Sentence Encoder (USE)~\cite{cer_2018_universal} are trained on datasets with billion samples, including sources from Wikipedia, web news, web question-answer pages and discussion forums.
 
Most of the related research on multimodal classification focuses on  cross-modality mapping of feature space. For instance, \cite{vo_2019_composing} embeds the image + text to an image space for retrieval. In this case, an image and text alternating/modifying  this image (alt-text, as referred in \cite{jia2021scaling}) is provided and the algorithm retrieves the most similar images according to such text description.
It is build on LSTM \cite{sak2014long} and  resnet \cite{he2016deep} for image and text encoding. Nonetheless, the descriptive ability of such deep learning models is not sufficient to classify abstract concepts in GCD.

The general approach to solve the proposed visual task is motivated by CLIP~\cite{radford_2021_learning} and USE \cite{cer_2018_universal}. Using advanced vision and natural language learning models, such as CLIP and USE, we attempted to create a methodology that is able to utilize higher level learning to correctly identify abstract concepts from images and text. 

Apart from the cover text and inside text given with the dataset, we propose to compute the captions of the greeting cards. These captions are the textual description of the visual features of the greeting cards. Various image captioning network \cite{clipcap,clipdiff} are helpful to compute the textual description of the greeting cards from visual features. 

Recently, prominent research work is contributed to text-to-image generation models. Diffusion models \cite{glide,imagen,dalle2} dominates the field of photo-realistic image generation from text prompts.  In our work, we propose to generate greeting cards images from pretrained text-to-image generation models.


Contributions: There are five main contributions of this work. 1) We introduce a multimodal dataset consisting of abstract concepts in greeting cards. The dataset offers challenges for the state-of-the art computer vision algorithms. 2) For this challenging data, we propose to embed the image and text into high level feature space using USE and CLIP architectures. The combination of CLIP/USE in our work gives us edge for strengthening CLIP image features with sentence level USE encoding. To the best of our knowledge, this is a first research to combine CLIP with USE encoding. 
3) However, simply combining these features representing different modalities is not optimal, since they contain redundant information, which may suppress their complementary information. Therefore, we propose to learn their optimal composition with respect to retrieval performance. 
4) Moreover, we propose to compute the captions of the image with pretrianed image caption model.
5) Finally, we demonstrate the greeting cards generation with the proposed greeting cards dataset.

\begin{figure*}
\begin{center}
\includegraphics[width = 6 in]{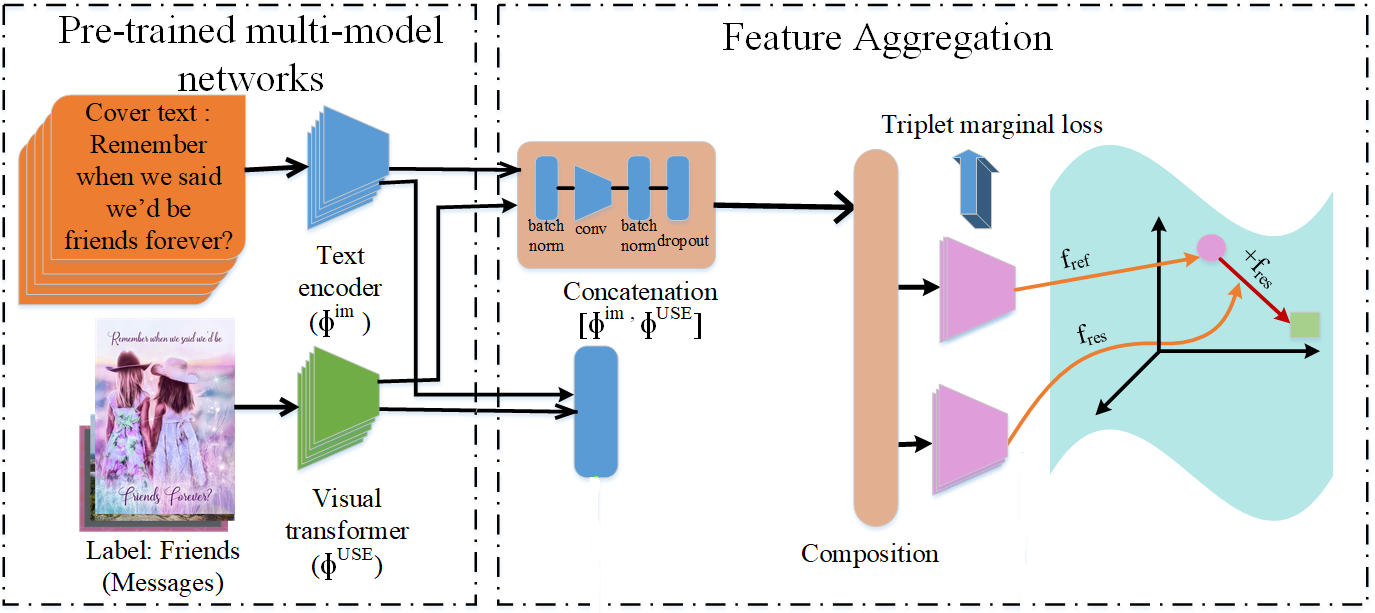}
   
\caption{The overall block diagram for CLIP and USE embedding and learning modified image features with text embedding}
\label{fig:BD}
\end{center} 
\end{figure*}

In this paper, we introduce a multi-modal greeting cards datasets. In our proposed work, we leverage pre-trained features from CLIP and USE for image and text encoder respectively. We also compute nearest neighbours of the node using text modified image features. The textual description of visual features of the greeting cards is discussed in section \ref{clip}, finally the image generation from pretrained modal is presented in section \ref{generation}. Overall, flow of the proposed work is shown in Fig.~\ref{fig:BD}. 

\section{Method}
\subsection{Feature Aggregation} \label{features}
We use a vision transformer based architecture CLIP \cite{radford_2021_learning} as image encoder and Universal Sentence Encoder \cite{cer_2018_universal} (USE)  as text encoder. The pre-trained weights of CLIP image encoder are loaded to vision transformer and pre-trained weights of USE V4 are loaded to text encoder to compute image $\phi_{im}$ and text $\phi_{USE}$ features for a given greeting card,
where CLIP is applied to the card cover image and USE to the card cover text.
Both image and text features are 512-D embedding. 


A simplest way is to concatenate 
the normalized features as
$F = [\phi_{im}, \phi_{USE}]$.
In this basic form of information aggregation, we do not know if we have redundant information conveyed by vision and text features. 
Since the redundant information may overpower the complementary information,
to counter this redundancy,
and increase the information content,
we propose to learn a single embedding using image and text features. 
This is one of the main contribution of our work.

First, we compose the image and text features by concatenating them with a single linear convolutional layer to merge the information of the image and text features,
followed by $ReLU$ and another linear convolutional filter.
\begin{equation}
f= W_{im 1}\left( {ReLU} \left(W_{lin} \left[\phi_{im}, \phi_{USE}\right]\right)\right),
\end{equation}
where $W_{lin}$ and $W_{im1}$ are the weights of a linear convolutional filters and \textit{f} tells us the joint information of $\phi_{im}$ and $\phi_{USE}$. 

Next, we intend to find the reference with respect to $\phi_{USE}$. So, we compute the dot product of \textit{f} and $\phi_{im}$ as follows:
\begin{equation}
f_{\text{ref}}\left(\phi_{im}, \phi_{USE}\right)=\sigma\left(f \odot \phi_{im}\right),
\end{equation}
where $\sigma$ is the sigmoid function and $\odot$ is element wise product. Then we learn a residual feature to $f_\text{ref}$.
\begin{equation}
f_{\text{r}} = W_{t 1}\left({RELU} \left(W_{lin} \left[\phi_{im}, \phi_{USE}\right]\right)\right)
\end{equation}
\begin{equation}
f_{\text{res}}\left(\phi_{im}, \phi_{USE}\right) = W_{t 2}\left({RELU} \left(f_{\text{r}} \right)\right)
\end{equation}
Here, $W_{lin}$ is a same convolutional filter from used eq (1), $W_{t 1}$ and $W_{t 1}$ are the two filters to learn residual feature $f_{\text{res}}$, they are called residual features since it is composed of the reference feature $f_{\text {ref}}$ to obtain our final embedding feature:
\begin{equation}
\phi^{im(R-text)}=w_{r} f_{\text {ref}}\left(\phi_{im}, \phi_{USE}\right)+w_{d} f_{\text {res}}\left(\phi_{im}, \phi_{USE}\right)
\end{equation}
$w_r$ and $w_d$ are learn-able weights. 
The image residual-text feature $\phi^{im(R-text)}$ represents a text complemented image embedding and can be viewed as a joint $\phi_{im}$ and $\phi_{USE}$ embedding.

For training the $im(R-text)$ network, we use the triplet marginal loss \cite{Yu_2019_ICCV} and mine the semi-hard triplet pairs to train the network end-to-end for learning $\phi^{im(R-text)}$ embedding from two complementary $\phi_{im}$ and $\phi_{USE}$ features. The loss function is given as follows:

\begin{equation}
\text{Loss} =\sum_{i=1}^{N}\left[\left\|f_{ia}-f_{ip}\right\|_{2}^{2}-\left\|f_{ia}-f_{in}\right\|_{2}^{2}+\alpha\right]_{+}
\end{equation}
where,
\begin{equation}
f = \phi^{im(R-text)}
\end{equation}

In our work, $\alpha$ = 0.2. Sub-scripts \textit{a}, \textit{p} and \textit{n} denote anchor, positive, and negative samples in a triplet for the $i^{th}$ sample in a batch. For $\phi^{im(R-text)}$, it is jointly learned by $\phi_{im}$ and $\phi_{USE}$ as explained above.

The intuitive explanation of the feature aggregation we discussed above is that the two features $\phi_{im}$ and $\phi_{USE}$ are conveying a combination of complementary and redundant information. Both $\phi_{im}$ and $\phi_{USE}$ focus on broad category tag information in GCD like which objects are present. 
For instance, if we consider greeting cards for \textit{Mother's day}, \textit{Father's day} and \textit{Grand parent's day} all of them may have similar visual objects as elderly man/women with child. But their cover text makes them different such as \textit{Happy Mother's Day to my Dear Mom},  \textit{You are OTTERLY terrific, Dad!}, and \textit{Happy Grandparent's Day Grandma and Grandpa} respectively. In this way, it can be noticed that the features $f_{\text {res}}$ would define a difference between the embedding assign to above mentioned three classes even tough $f_{\text{ref}}$ could be the same given the similar visual content of their cards.
\begin{figure*}[!ht]
\begin{center}
\includegraphics[scale = 0.55]{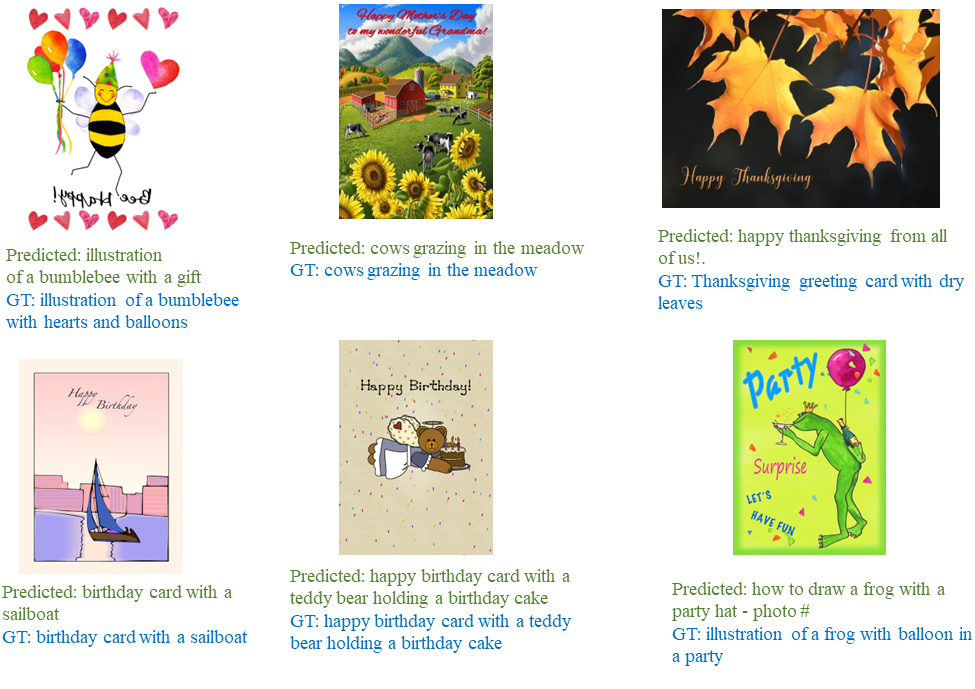}
   
\caption{The predicted caption using pretrained clipcap model \cite{clipcap}}
\label{fig:clipcap}
\end{center} 
\end{figure*}
\begin{figure*}[!ht]
    \centering
\includegraphics[scale=1]{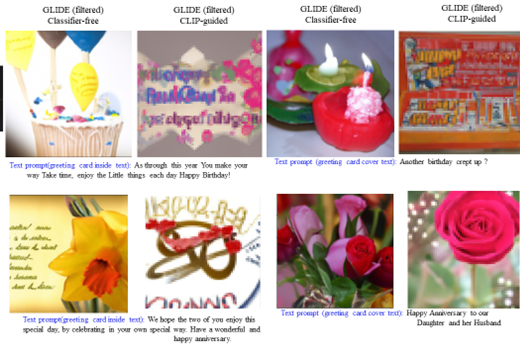}
\caption{The greeting cards generated from GLIDE \cite{glide} pretrained model using classifier free and CLIP guided model} 
\label{fig:GLIDE_samples}
\end{figure*}
Fusion of image and text features have frequently been used in previous research \cite{vo_2019_composing, anwaar2021compositional, wu2021fashion}. To the best of our knowledge, all of them are for cross-modal retrieval, such as image and text pair are given to retrieve the most similar images. The closest to our work is \cite{vo_2019_composing}. However, their architecture relies on resnet and LSTM layers for feature computation. In our case, resnet and LSTM gives lower performance than pretrained visual and text features as shown in section \ref{results}. 
In \cite{vo_2019_composing}, they use more complex convolutional filters because they have spatial features from last convolutional layer from resnet and LSTM, but we use linear filters on pretrained features. 


In our work, $\phi^{im(R-text)}$ features help us to make a better graph structured data, since they improve the neighbourhood, which is important of better feature aggregation with GNN. To the best of our knowledge, this is a first work to construct a graph using fused multi-modal (image, text) features $\phi^{im(R-text)}$. Different graph construction schemes using pre-trained features and learned $\phi^{im(R-text)}$ features is described in the next section. 




\subsection{Image caption prediction} \label{clip}
The captions for the GCD dataset are computed with the pre-trained clip captioning model (clipcap) \cite{clipcap}. The clipcap model in \cite{clipcap} use CLIP encoding and fine-tunes a language model to generate the image captions. The predicted captions for the proposed GCD dataset using clipcap pre-trained model is shown in Fig. \ref{fig:clipcap}. The clipcap model \cite{clipcap}  generates meaningful captions for GCD datasets.

\subsection{Text-to-image generation} \label{generation}
The proposed dataset is also useful for generating greeting cards images using text-to-image generation model. In this work, we generated the sample greeting cards images using GLIDE pretrained model \cite{glide}. The examples shown in Fig. \ref{fig:GLIDE_samples} are generated from the GLIDE model (filtered) for the given text prompts. We used two pre-trained GLIDE model, one is classifier free model and the other is CLIP guided model. We used cover text and inside text as text prompts to generate image samples separately.




\section{Experiments}\label{exp}

\subsection{Datasets}

Our dataset GCD comprises of 3700+ unique images and text for greeting cards. For experimentation, we randomly split it into 80\% for training and 20\% for testing for each category, i.e., Holidays, Messages and Special Occasions. In each category, there are several subcategories as listed inFig. \ref{fig:num_images}. We evaluate the classification accuracy of each subcategory and average the classification accuracy of all the subcategories to compute the classification accuracy of a category as shown in Table \ref{tab:pretrianed}. 


\subsection{Results for pretrained networks} \label{results}

\begin{table}[]
\begin{center}
\begin{tabular}{c|c|c|c}
\hline
Method & $\phi^{im}$    & $\phi^{USE}$   & [$\phi^{im}$, $\phi^{USE}$]  \\ \hline
Holidays               & 70.12 & 75.3 & 77.7            \\ \hline
\begin{tabular}[c]{@{}c@{}}Special   Occasions\end{tabular} & 47.17 & 54.7 & 58.5           \\ \hline
Messages                                                      & 48.79 & 63.57 & 66           \\ \hline
Average          & 55.36                                                        &64.5                                                       & 67.4                                                                    \\ \hline
\end{tabular}
\caption{Classification accuracy GCD of kNN for k=20. $\phi^{im}$ stands for embedding from CLIP visual transformer, $\phi^{USE}$ stands for embedding for USE sentence encoder and [$\phi^{im}$, $\phi^{USE}$] stands for the normalized concatenation of $\phi^{im}$ and $\phi^{USE}$}
\label{tab:pretrianed}
\end{center}
\end{table}


First, we establish a baseline for the GCD. The baseline embedding is computed on pretrained  CLIP visual transformer and USE. For this purpose, we download the pretrained network weights from CLIP and USE V4 to compute the feature vector for images and text on GCD. The results on these embeddings for classification using a knn-classifier 
are shown in Table ~\ref{tab:pretrianed}. The embedding from CLIP visual transformer and USE encoder are denoted by $\phi^{im}$ and $\phi^{USE}$ respectively. [$\phi^{im}$,$\phi^{USE}$] denotes the normalized concatenation of $\phi^{im}$ and $\phi^{USE}$.
The concatenation of both features achieves better performance than separate embeddings. The CLIP feature alone performs poorly since CLIP is not good enough to capture sentiments in cover text. Especially for categories like \textit{"Messages"} in GCD which have more emotions expressed as words than visual content. The best performance for the pre-trianed CLIP and USE embedding is obtained by concatenating the normalized embeddings [$\phi^{im}$,$\phi^{USE}$]. Just by concatenating, classification accuracy improves significantly in comparison to CLIP visual encoder ($\phi^{im}$).


\begin{table}[]
\begin{center}
\begin{tabular}{c|c|c|c}
\hline
Data category     & \begin{tabular}[c]{@{}c@{}} TIRG \\ \cite{vo_2019_composing}\end{tabular} &  \begin{tabular}[c]{@{}c@{}} CLIP \\ 
\cite{radford_2021_learning}\end{tabular} & \begin{tabular}[c]{@{}c@{}} $\phi^{im(R-text)}$ \\ \end{tabular} \\ \hline
Holidays          & 79.06  & 80.44 & \textbf{85.12}   \\ \hline
Special Occasions & 71.7   & \textbf{81.13}   & 79.25 \\ \hline
Messages          & 63.63  & 66.56  & \textbf{73.05}  \\ \hline
Average          & 71.46 & 76.04   & \textbf{79.14} \\ \hline
\end{tabular}
\caption{The (cross entropy) classification accuracy. TIRG and CLIP are trained end-to-end on GCD. Both the networks utilize image and text features.}
\label{tab:CNN_trained}
\end{center}
\end{table}

\subsection{Results for models trained on GCD}
Apart from classifying images in GCD using pretrained embeddings and KNN classification, we also trained CLIP \cite{radford_2021_learning} and TIRG \cite{vo_2019_composing} network in an end-to-end manner. For training the CLIP visual and text encoders, we concatenate the output of visual and text transformers and apply a cross-entropy loss function after passing them through a fully connected layer. The classification accuracy is listed in Table \ref{tab:CNN_trained} (second column). 

Moreover, on average, the classification accuracy for pretrained $CLIP$'s text encoder and $USE$ encoder is 49.4\% and  70.32\% respectively. 
Since the CLIP text encoder is trained for word-based embeddings, the advantage of sentence encoding is missing in this architecture. For this reason we prefer USE \cite{cer_2018_universal} over CLIP text encoder in our approach.

Similarly, we train and evaluate a CNN based network TIRG \cite{vo_2019_composing} on GCD. It consists of resnet and LSTM backbone for feature extraction and $\phi^{im(R-text)}$ layer for feature aggregation. Herer $im$ is computed by resnet and $R-text$ is computed by LSTM. The results are listed in Table \ref{tab:CNN_trained} (first column). For CNN based network, we finally trained and evaluated integrate the CLIP and USE features with $\phi^{im(R-text)}$ layer for feature aggregation ~\cite{radford_2021_learning, vo_2019_composing} as shown in Table \ref{tab:CNN_trained} (third column). It improves the classification accuracy from 76.04\% to 79.14\% as shown in Table \ref{tab:CNN_trained}. 
Hence, on average the proposed $(\phi^{im(R-text)})$ shows the best accuracy amongst models trained on GCD.

\begin{figure*}[!ht]
    \centering
\includegraphics[scale=1]{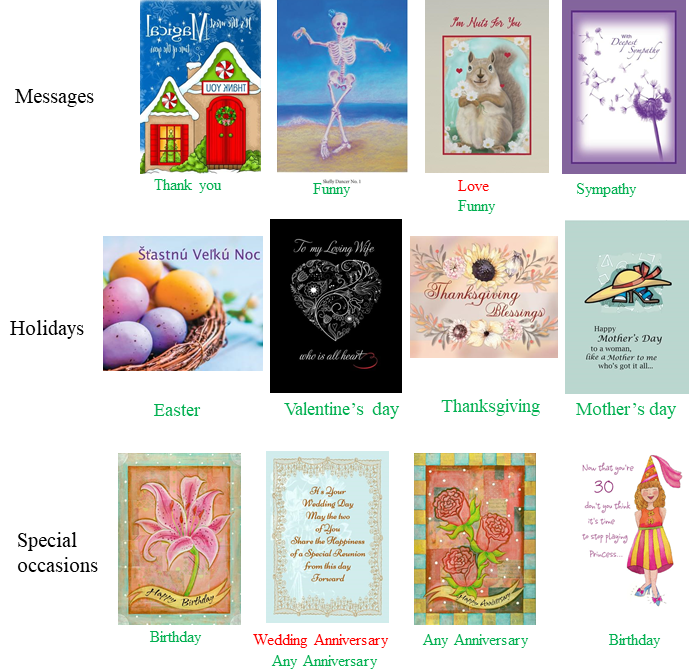}
\caption{The qualitative results for classification on example images from Holidays, Special occasions and Messages categories. Below each image, the predicted label is shown in \textcolor{green}{green} if the predicted and actual labels are same. If the predicted and actual labels are not same, then the correct label is shown in \textcolor{red}{red} and predicted label in \textcolor{green}{green}.} 
\label{fig:sample}
\end{figure*}
%

\vspace{\baselineskip}



\section{Conclusion}
In our work, we introduce a novel dataset for predicting a labels for greeting cards images. The propose method fuses text and image representations to construct the baseline for the classification of the multi-modal greeting cards dataset. The proposed dataset could be use to compute the textual description of greeting cards from pre-trained image captioning modals. The textual description of the greeting cards would eventually be used to generate greeting cards from text prompts by leveraging pretrained text-to-image generation networks. The novel dataset introduced in our paper with abstract images will contribute in moving to new paradigm of pre-trained modals in multi-modal learning for abstract images.

{\small
\bibliographystyle{ieee_fullname}
\bibliography{egpaper}

\begin{thebibliography}{10}\itemsep=-1pt

\bibitem{signedcards}
signedcards.com.

\bibitem{anwaar2021compositional}
Muhammad~Umer Anwaar, Egor Labintcev, and Martin Kleinsteuber.
\newblock Compositional learning of image-text query for image retrieval.
\newblock In {\em Proceedings of the IEEE/CVF Winter Conference on Applications
  of Computer Vision}, pages 1140--1149, 2021.

\bibitem{cer_2018_universal}
Daniel Cer, Yinfei Yang, Sheng-yi Kong, Nan Hua, Nicole Limtiaco, Rhomni~St
  John, Noah Constant, Mario Guajardo-Cespedes, Steve Yuan, Chris Tar,
  Yun-Hsuan Sung, Brian Strope, and Ray Kurzweil.
\newblock Universal sentence encoder, 2018.

\bibitem{changpinyo2021cc12m}
Soravit Changpinyo, Piyush Sharma, Nan Ding, and Radu Soricut.
\newblock {Conceptual 12M}: Pushing web-scale image-text pre-training to
  recognize long-tail visual concepts.
\newblock In {\em CVPR}, 2021.

\bibitem{he2016deep}
Kaiming He, Xiangyu Zhang, Shaoqing Ren, and Jian Sun.
\newblock Deep residual learning for image recognition.
\newblock In {\em Proceedings of the IEEE conference on computer vision and
  pattern recognition}, pages 770--778, 2016.

\bibitem{jia2021scaling}
Chao Jia, Yinfei Yang, Ye Xia, Yi-Ting Chen, Zarana Parekh, Hieu Pham, Quoc~V
  Le, Yunhsuan Sung, Zhen Li, and Tom Duerig.
\newblock Scaling up visual and vision-language representation learning with
  noisy text supervision.
\newblock {\em arXiv preprint arXiv:2102.05918}, 2021.

\bibitem{krishna2017visual}
Ranjay Krishna, Yuke Zhu, Oliver Groth, Justin Johnson, Kenji Hata, Joshua
  Kravitz, Stephanie Chen, Yannis Kalantidis, Li-Jia Li, David~A Shamma, et~al.
\newblock Visual genome: Connecting language and vision using crowdsourced
  dense image annotations.
\newblock {\em International journal of computer vision}, 123(1):32--73, 2017.

\bibitem{lin2014microsoft}
Tsung-Yi Lin, Michael Maire, Serge Belongie, James Hays, Pietro Perona, Deva
  Ramanan, Piotr Doll{\'a}r, and C~Lawrence Zitnick.
\newblock Microsoft coco: Common objects in context.
\newblock In {\em European conference on computer vision}, pages 740--755.
  Springer, 2014.

\bibitem{clipcap}
Ron Mokady, Amir Hertz, and Amit~H Bermano.
\newblock Clipcap: Clip prefix for image captioning.
\newblock {\em arXiv preprint arXiv:2111.09734}, 2021.

\bibitem{glide}
Alex Nichol, Prafulla Dhariwal, Aditya Ramesh, Pranav Shyam, Pamela Mishkin,
  Bob McGrew, Ilya Sutskever, and Mark Chen.
\newblock Glide: Towards photorealistic image generation and editing with
  text-guided diffusion models.
\newblock {\em arXiv preprint arXiv:2112.10741}, 2021.

\bibitem{ordonez2011im2text}
Vicente Ordonez, Girish Kulkarni, and Tamara Berg.
\newblock Im2text: Describing images using 1 million captioned photographs.
\newblock {\em Advances in neural information processing systems},
  24:1143--1151, 2011.

\bibitem{plummer2015flickr30k}
Bryan~A Plummer, Liwei Wang, Chris~M Cervantes, Juan~C Caicedo, Julia
  Hockenmaier, and Svetlana Lazebnik.
\newblock Flickr30k entities: Collecting region-to-phrase correspondences for
  richer image-to-sentence models.
\newblock In {\em Proceedings of the IEEE international conference on computer
  vision}, pages 2641--2649, 2015.

\bibitem{radford_2021_learning}
Alec Radford, Jong~Wook Kim, Chris Hallacy, Aditya Ramesh, Gabriel Goh,
  Sandhini Agarwal, Girish Sastry, Amanda Askell, Pamela Mishkin, Jack Clark,
  Gretchen Krueger, and Ilya Sutskever.
\newblock Learning transferable visual models from natural language
  supervision, 2021.

\bibitem{dalle2}
Aditya Ramesh, Prafulla Dhariwal, Alex Nichol, Casey Chu, and Mark Chen.
\newblock Hierarchical text-conditional image generation with clip latents.
\newblock {\em arXiv preprint arXiv:2204.06125}, 2022.

\bibitem{imagen}
Chitwan Saharia, William Chan, Saurabh Saxena, Lala Li, Jay Whang, Emily
  Denton, Seyed Kamyar~Seyed Ghasemipour, Burcu~Karagol Ayan, S~Sara Mahdavi,
  Rapha~Gontijo Lopes, et~al.
\newblock Photorealistic text-to-image diffusion models with deep language
  understanding.
\newblock {\em arXiv preprint arXiv:2205.11487}, 2022.

\bibitem{sak2014long}
Hasim Sak, Andrew~W Senior, and Fran{\c{c}}oise Beaufays.
\newblock Long short-term memory recurrent neural network architectures for
  large scale acoustic modeling.
\newblock 2014.

\bibitem{srinivasan2021wit}
Krishna Srinivasan, Karthik Raman, Jiecao Chen, Michael Bendersky, and Marc
  Najork.
\newblock Wit: Wikipedia-based image text dataset for multimodal multilingual
  machine learning.
\newblock {\em arXiv preprint arXiv:2103.01913}, 2021.

\bibitem{vo_2019_composing}
Nam Vo, Lu Jiang, Chen Sun, Kevin Murphy, Li-Jia Li, Li Fei-Fei, and James
  Hays.
\newblock Composing text and image for image retrieval - an empirical odyssey.
\newblock {\em Thecvf.com}, pages 6439--6448, 2019.

\bibitem{wu2021fashion}
Hui Wu, Yupeng Gao, Xiaoxiao Guo, Ziad Al-Halah, Steven Rennie, Kristen
  Grauman, and Rogerio Feris.
\newblock Fashion iq: A new dataset towards retrieving images by natural
  language feedback.
\newblock In {\em Proceedings of the IEEE/CVF Conference on Computer Vision and
  Pattern Recognition}, pages 11307--11317, 2021.

\bibitem{clipdiff}
Shitong Xu.
\newblock Clip-diffusion-lm: Apply diffusion model on image captioning.
\newblock {\em arXiv preprint arXiv:2210.04559}, 2022.

\bibitem{Yu_2019_ICCV}
Baosheng Yu and Dacheng Tao.
\newblock Deep metric learning with tuplet margin loss.
\newblock In {\em Proceedings of the IEEE/CVF International Conference on
  Computer Vision (ICCV)}, October 2019.

\end{thebibliography}
}

\end{document}